\documentclass[runningheads]{llncs}
\usepackage[table]{xcolor}
\usepackage{stmaryrd}
\usepackage{multirow}
\usepackage{tabularx}
\usepackage{booktabs}
\usepackage{bbm}
\usepackage{breqn}
\usepackage{mathtools}
\usepackage{bm}
\usepackage{amssymb}
\usepackage{amsmath}

\usepackage{tikz}
\usepackage{subfigure}
\usetikzlibrary{backgrounds,automata}
\usetikzlibrary{bayesnet}
\usepackage{xspace}

\newcommand\Ttwo{\text{T}\textsubscript{2}}
\newcommand\iUS{\text{iUS}}

\DeclareMathOperator{\KL}{KL}

\usepackage{xcolor,colortbl}

\definecolor{Gray}{gray}{0.85}
\definecolor{LightGray}{gray}{0.95}
\newcolumntype{a}{>{\columncolor{Gray}}c}

\usepackage{hyperref}

\begin{document}
\title{Unified Brain MR-Ultrasound Synthesis using Multi-Modal Hierarchical Representations}
\titlerunning{Unified Synthesis using Multi-Modal Hierarchical Representations}
%
\author{Reuben Dorent\inst{1} \and
Nazim Haouchine\inst{1} \and
Fryderyk Kogl\inst{1} \and 
Samuel Joutard\inst{2} \and 
Parikshit Juvekar\inst{1}\and
Erickson Torio\inst{1}\and
Alexandra Golby\inst{1}\and
Sebastien Ourselin\inst{2}\and
Sarah Frisken\inst{1}\and
Tom Vercauteren\inst{2}\and
Tina Kapur\inst{1}\and
William M. Wells III\inst{1,3}}
%
\institute{Harvard Medical School, Brigham and Women's Hospital, Boston, MA, USA \and King’s College London, London, United Kingdom \and Massachusetts Institute of Technology, Cambridge, MA, USA
\email{rdorent@bwh.harvard.edu}
}
\authorrunning{Dorent et al}
%

%
\maketitle              
\begin{abstract}
We introduce MHVAE, a deep hierarchical variational auto-encoder (VAE) that synthesizes missing images from various modalities. Extending multi-modal VAEs with a hierarchical latent structure, we introduce a probabilistic formulation for fusing multi-modal images in a common latent representation while having the flexibility to handle incomplete image sets as input. Moreover, adversarial learning is employed to generate sharper images. Extensive experiments are performed on the challenging problem of joint intra-operative ultrasound (iUS) and Magnetic Resonance (MR) synthesis. Our model outperformed multi-modal VAEs, conditional GANs, and the current state-of-the-art unified method (ResViT) for synthesizing missing images, demonstrating the advantage of using a hierarchical latent representation and a principled probabilistic fusion operation. Our code is publicly available\footnote{\url{https://github.com/ReubenDo/MHVAE}}.
\end{abstract}

\section{Introduction}
Medical imaging is essential during diagnosis, surgical planning, surgical guidance, and follow-up for treating brain pathology. Images from multiple modalities 
are typically acquired to distinguish clinical targets from surrounding tissues. For example, intra-operative ultrasound (\iUS) imaging and Magnetic Resonance Imaging (MRI) capture complementary characteristics of brain tissues that can be used to guide brain tumor resection. However, as noted in~\cite{wu2018multimodal}, multi-modal data is \textit{expensive} and \textit{sparse}, typically leading to incomplete sets of images. For example, the prohibitive cost of intra-operative MRI (iMRI) scanners often hampers the acquisition of iMRI during surgical procedures. Conversely, iUS is an affordable tool but has been perceived as difficult to read compared to iMRI \cite{dixon2022intraoperative}. Consequently, there is growing interest in synthesizing missing images from a subset of available images for enhanced visualization and clinical training.

Medical image synthesis aims to predict missing images given available images. Deep-learning based methods have reached the highest level of performance~\cite{wang2021review}, including conditional generative adversarial (GAN) models~\cite{isola2017image,park2019semantic,donnez2021realistic,9174648} and conditional variational auto-encoders \cite{chartsias2019disentangled}. However, a key limitation of these techniques is that they must be trained for each subset of available images. 

\begin{figure}[tb!]
    \centering
    \subfigure[VAE]{\includegraphics[width=0.18\textwidth]{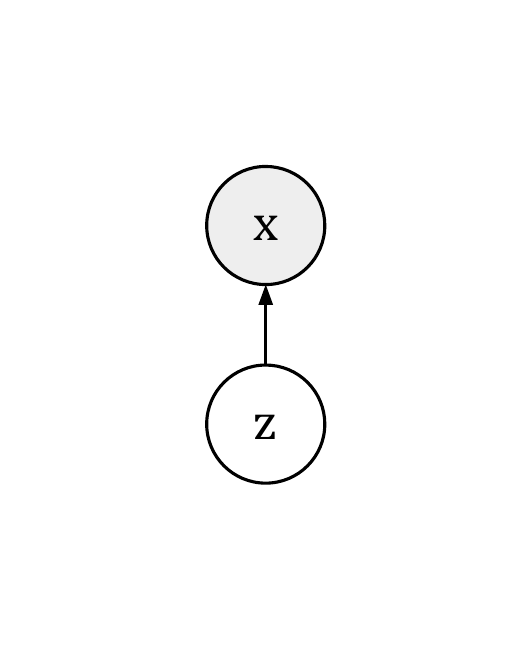}} 
    \subfigure[HVAE]{\includegraphics[width=0.18\textwidth]{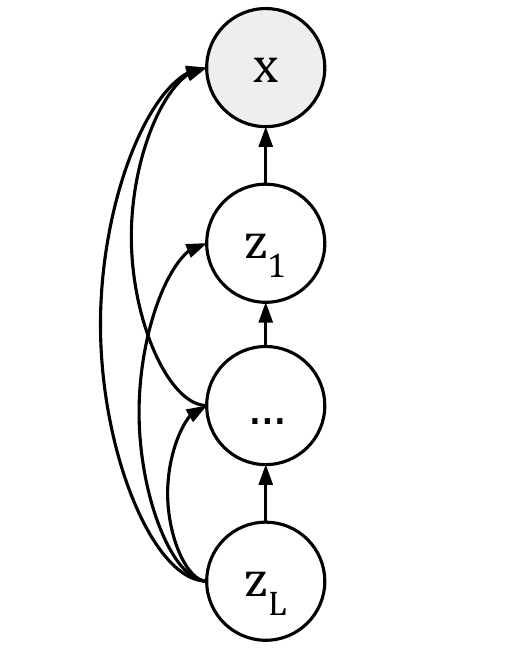}} 
    \subfigure[MVAE]{\includegraphics[width=0.18\textwidth]{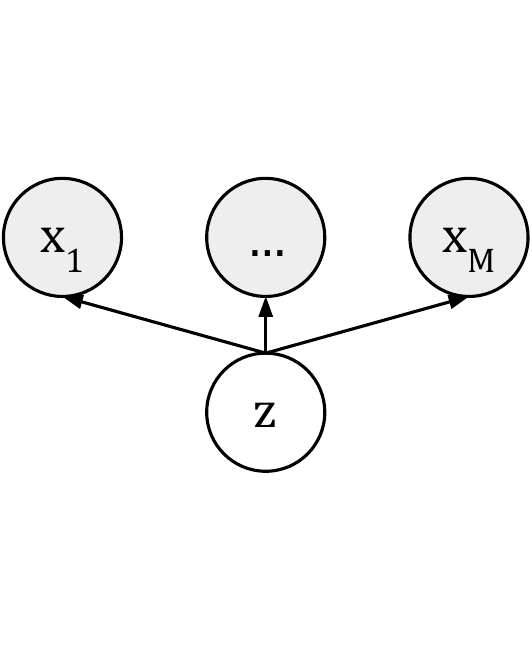}}
    \subfigure[Ours]{\includegraphics[width=0.18\textwidth]{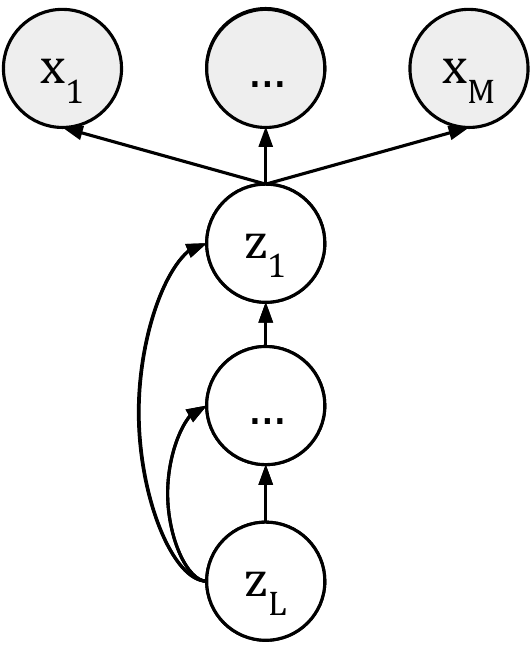}}
    \caption{Graphical models of: (a) variational auto-encoder (VAE); (b) hierarchical VAE (HVAE); (c) Multi-modal VAE (MVAE); (d) Multi-Modal Hiearchical VAE (Ours). }
    \label{fig:graphical_models}
\end{figure}

To tackle this challenge, unified approaches have been proposed. These approaches are designed to have the flexibility to handle incomplete image sets as input, improving practicality as only one network is used for generating missing images. To handle partial inputs, some studies proposed to fill missing images with arbitrary values \cite{Sharma20,resvit,li2019diamondgan,lee2019collagan}. Alternatively, other work aim at creating a common feature space that encodes shared information from different modalities. Feature representations are extracted independently for each modality. Then, arithmetic operations (e.g., mean \cite{hemis,pimms,dorent2021learning},  max \cite{chartsias2017multimodal} or a combination of sum, product and max \cite{zhou2020hi}) are used to fuse these feature representations. However, these operations do not force the network to learn a shared latent representation of multi-modal data and lack theoretical foundations. In contrast, Multi-modal Variational Auto-Encoders (MVAEs) provide a principled probabilistic fusion operation to create a common representation space \cite{wu2018multimodal,dorent2019hetero}.
In MVAEs, the common representation space is low-dimensional (e.g., $\mathbb{R}^{256}$), which usually leads to blurry synthetic images. In contrast, hierarchical VAEs (HVAEs)~\cite{vahdat2020nvae,ranganath2016hierarchical,maaloe2019biva,sonderby2016ladder} allow for learning complex latent representations by using a hierarchical latent structure, where the coarsest latent variable ($z_L$) represents global features, as in MVAEs, while the finer variables capture local characteristics. However, HVAEs have not yet been extended to multi-modal settings to synthesize missing images.

In this work, we introduce Multi-Modal Hierarchical Latent Representation VAE (MHVAE), the first multi-modal VAE approach with a hierarchical latent representation for unified medical image synthesis. Our contribution is four-fold. First, we integrate a hierarchical latent representation into the multi-modal variational setting to improve the expressiveness of the model.
Second, we propose a principled fusion operation derived from a probabilistic formulation to support missing modalities, thereby enabling image synthesis. 
Third, adversarial learning is employed to generate realistic image synthesis. 
Finally, experiments on the challenging problem of iUS and MR synthesis demonstrate the effectiveness of the proposed approach, enabling the synthesis of high-quality images while establishing a mathematically grounded formulation for unified image synthesis and outperforming non-unified  GAN-based approaches and the state-of-the-art method for unified multi-modal medical image synthesis.

\section{Background}
\subsubsection{Variational Auto-Encoders (VAEs).}
The goal of VAEs~\cite{kingma2013auto} is to train a generative model in the form of $p(x, z)=p(z)p(x|z)$ where $p(z)$ is a prior distribution (e.g. isotropic Normal distribution) over latent variables $z\in\mathbb{R}^{H}$  and where $p_{\theta}(x|z)$ is a decoder parameterized by $\theta$ that reconstructs data $x\in\mathbb{R}^{N}$ given $z$. The latent space dimension $H$ is typically much lower than the image space dimension $N$, i.e. $H\ll N$. The training goal with respect to $\theta$ is to maximize the marginal likelihood of the data $p_{\theta}(x)$ (the ``evidence''); however since the true posterior $p_{\theta}(z|x)$ is in general intractable, the variational evidence lower bound (ELBO) is instead optimized. The ELBO $\mathcal{L}_{\text{VAE}}(x; \theta, \phi)$ is defined by introducing an approximate posterior $q_{\phi}(z|x)$ with parameters $\phi$:
\begin{equation}
    \label{eqq:elbo_vae}
    \mathcal{L}_{\text{VAE}}(x; \theta, \phi)  \coloneqq \mathbb{E}_{q_{\phi}(z|x)}[\log(p_{\theta}(x|z))] - \KL[q_{\phi}(z|x)||p(z)] \ ,
\end{equation}
where $\KL[q||p]$ is the Kullback-Leibler divergence between distributions $q$ and $p$.

\subsubsection{Multi-modal Variational Auto-Encoders (MVAE)}
Multi-modal VAEs~\cite{wu2018multimodal,dorent2019hetero,shi2019variational} introduced a principled probabilistic formulation to support missing data at training and inference time. Multi-modal VAEs assume that $M$ paired images $x=(x_1,...,x_M) \in \mathbb{R}^{M \times N}$ are conditionally independent given a shared representation $z$ as higlighted in Fig.\ref{fig:graphical_models}, i.e. $p_{\theta}(x|z)=\prod_{i=1}^{M}p(x_i|z)$.

Instead of training one single variational  network $q_{\phi}(z|x)$ that requires all images to be presented at all times, MVAEs factorize the approximate posterior as a combination of unimodal variational posteriors $(q_{\phi}(z|x_i))_{i=1}^{M}$. Given any subset of modalities $\pi\subseteq\{1,...,M\}$, MVAEs have the flexibility to approximate the $\pi$-marginal posteriors $p(z|(x_{i})_{{i\in\pi}})$ using the $|\pi|$ unimodal variational posteriors $(q_{\phi}(z|x_i))_{i\in \pi}$. MVAE~\cite{wu2018multimodal} and U-HVED~\cite{dorent2019hetero} factorize the $\pi$-marginal variational posterior as a product-of-experts (PoE), i.e.: 
\begin{equation}
    q^{\text{PoE}}_{\phi}(z|x_{\pi})=p(z)\prod_{i\in\pi}q_{\phi}(z|x_i) \ .
\end{equation}

\section{Methods}
In this paper, we propose a deep multi-modal hierarchical VAE called MHVAE that synthesizes missing images from available images. MHVAE’s design focuses on tackling three challenges: (i) improving expressiveness of VAEs and MVAEs using a hierarchical latent representation; (ii) parametrizing the variational posterior to handle missing modalities; (iii) synthesizing realistic images.

\subsection{Hierarchical latent 
representation} 
Let $x=(x_i)_{i=1}^M \in \mathbb{R}^{M \times N}$ be a complete set of paired (i.e. co-registered) images of different modalities where $M$ is the total number of image modalities and $N$ the number of pixels (e.g. $M=2$ for \Ttwo \ MRI and iUS synthesis).
The images $x_i$ are assumed to be conditionally independent given a latent variable $z$. Then, the conditional distribution $p_{\theta}(x|z)$ parameterized by $\theta$ can be written as:
\begin{equation}
    p_{\theta}(x|z)=\prod_{i=1}^{M}p_{\theta}(x_i|z) \ .
\end{equation}

Given that VAEs and MVAEs typically produce blurry images, we propose to use a hierarchical representation of the latent variable $z$ to increase the expressiveness the model as in HVAEs~\cite{vahdat2020nvae,ranganath2016hierarchical,maaloe2019biva,sonderby2016ladder}. Specifically, the latent variable $z$ is partitioned into disjoint groups, as shown in Fig.\ref{fig:graphical_models} i.e. $z=\{z_1,...z_L\}$, where $L$ is the number of groups. The prior $p(z)$ is  then represented by:
\begin{equation}
    p_{\theta}(z)=p(z_L)\prod_{l=1}^{L-1}p_{\theta_{l}}(z_l|z_{>l}) \ ,
\end{equation}
where $p(z_L)=\mathcal{N}(z_L; 0, I)$ is an isotropic Normal prior distribution and the conditional prior distributions $p_{\theta_{l}}(z_l|z_{>l})$ are factorized Normal distributions with diagonal covariance parameterized using neural networks, i.e. $p_{\theta_{l}}(z_l|z_{>l})=\mathcal{N}(z_l; \mu_{\theta_{l}}(z_{>l}), D_{\theta_{l}}(z_{>l}))$. Note that the dimension of the finest latent variable $z_1\in\mathbb{R}^{H_1}$ is  similar to number of pixels, i.e. $H_1=\mathcal{O}(N)$ and the dimension of the latent representation exponentially decreases with the depth, i.e. $H_L \ll H_1$.

Reusing  Eq.~\ref{eqq:elbo_vae}, the evidence $\log\left(p_{\theta}\left(x\right)\right)$ is lower-bounded by the tractable variational ELBO  $\mathcal{L}^{\text{ELBO}}_{\text{MHVAE}}(x; \theta, \phi)$:
\begin{equation}
\label{eq:elbo_MHVAE_all}
\begin{split}
\mathcal{L}^{\text{ELBO}}_{\text{MHVAE}}(x; \theta, \phi)= & \sum_{i=1}^{M}\mathbb{E}_{q_{\phi}(z|x)}[\log(p_{\theta}(x_i|z))] - \KL\left[q_{\phi}(z_L|x)||p(z_L)\right] \\
 {}& \quad - \sum_{l=1}^{L-1} \mathbb{E}_{q_{\phi}(z_{>l}|x)}\left[ \KL[q_{\phi}(z_{l}|x,z_{>l})||p_{\theta}(z_{l}|z_{>l})] \right] 
\end{split}
\end{equation}
where $q_{\phi}(z|x) = \prod_{l=1}^{L}q_{\phi}(z_{l}|x,z_{>l})$ is a variational posterior that approximates the intractable true posterior $p_{\theta}(z|x)$.

\subsection{Variational posterior's parametrization for incomplete inputs}
To synthesize missing images, the variational posterior $(q_{\phi}(z_{l}|x,z_{>l}))_{l=1}^{L}$ should handle missing images. We propose to parameterize  it as a combination of unimodal variational posteriors. Similarly to MVAEs, for any set $\pi\subseteq\{1,...,M\}$ of images, the conditional posterior distribution at the coarsest level $L$ is expressed 
\begin{equation}
    q^{\text{PoE}}_{\phi_{L}}(z_{L}|x_{\pi})=p(z_{L})\prod_{i\in\pi}q_{\phi^{i}_{L}}(z|x_i) \ .
\end{equation}
where $p(z_L)=\mathcal{N}(z_L; 0, I)$ is an isotropic Normal prior distribution and $q_{\phi_{L}}(z|x_i)$ is a Normal distribution with diagonal covariance parameterized using CNNs.

For the other levels $l\in\{1,..,L-1\}$, we similarly propose to express the conditional variational posterior distributions as a product-of-experts:
\begin{equation}
    q^{\text{PoE}}_{\phi_l,\theta_l}(z_{l}|x_{\pi},z_{>l})=p_{\theta_l}(z_l|z_{>l})\prod_{i\in\pi} q_{\phi^{i}_{l}}(z_l|x_i,z_{>l})
\end{equation}
where $q_{\phi^{i}_{l}}(z_l|x_i,z_{>l})$ is a Normal distribution with diagonal covariance parameterized using CNNs, i.e. $q_{\phi^{i}_{l}}(z_l|x_i,z_{>l})=\mathcal{N}(z_l; \mu_{\phi^{i}_{l}}(x_i,z_{>l}); D_{\phi^{i}_{l}}(x_i,z_{>l}))$.

This formulation allows for a principled operation to fuse
content information from available images while having the flexibility to handle missing ones. Indeed, at each level $l\in\{1,...,L\}$, the conditional variational distributions $q^{\text{PoE}}_{\phi_l,\theta_l}(z_l|x_{\pi},z_{>l})$ are Normal distributions with mean $\mu_{\phi_{l},\theta_{l}}(x_{\pi},z_{>l})$ and diagonal covariance $D_{\phi_{l},\theta_{l}}(x_{\pi},z_{>l})$ expressed in closed-form solution~\cite{hernandez2010balancing} as:
\begin{equation*}
    \begin{cases}
        D_{\phi_{l},\theta_{l}}(x_{\pi},z_{>l})=\left(D^{-1}_{\theta_{l}}(z_{>l}) + \sum_{i\in\pi} D^{-1}_{\phi^{i}_{l}}(x_{i},z_{>l})\right)^{-1} \\
        \mu_{\phi_{l},\theta_{l}}(x_{\pi},z_{>l}) = D^{-1}_{\phi_{l},\theta_{l}}(x_{\pi},z_{>l}) \left(
    D^{-1}_{\theta_{l}}(z_{>l})\mu_{\theta_{l}}(z_{>l}) +
    \sum\limits_{i\in\pi} D^{-1}_{\phi^{i}_{l}}(x_{i},z_{>l})\mu_{\phi^{i}_{l}}(x_{i},z_{>l}) \right) \\
    \end{cases}
    \label{product_Normal_formula}
\end{equation*} 
with $D_{\theta_{L}}(z_{>L})=I$ and $\mu_{\theta_{L}}(z_{>L})=0$.

\subsection{Optimization strategy for image synthesis}\label{sec:optimization}
The joint reconstruction and synthesis optimization goal is to maximize the expected evidence $\mathbb{E}_{x\sim p_{\text{data}}}\left[\log(p(x))\right]$. As the ELBO defined in Eq.~\ref{eq:elbo_MHVAE_all} is valid for any approximate distribution $q$, the evidence, $\log(p_{\theta}(x))$, is in particular lower-bounded by the following subset-specific ELBO for any subset of images $\pi$:
\begin{equation}
\label{eq:lowerbound_partial}
\begin{split} 
\mathcal{L}^{\text{ELBO}}_{\text{MAVAE}}(x_{\pi}; \theta, \phi)= \underbrace{\sum_{i=1}^{M}\mathbb{E}_{q_{\phi}(z|x_{\pi})}[\log(p_{\theta}(x_i|z_{1}))]}_{\text{reconstruction}} - \KL\left[q_{\phi_L}(z_L|x_{\pi})||p(z_L)\right]  \\
 \quad - \sum_{l=1}^{L-1} \mathbb{E}_{q_{\phi_l,\theta_l}(z_{>l}|x_{\pi})}\left[ \KL[q_{\phi_l,\theta_l}(z_{l}|x_{\pi},z_{>l})||p_{\theta_l}(z_{l}|z_{>l})] \right] \ .
\end{split}
\end{equation}
Hence, the expected evidence $\mathbb{E}_{x\sim p_{\text{data}}}\left[\log(p(x))\right]$ is lower-bounded by the average of the subset-specific ELBO, i.e.:
\begin{equation}
\label{eq:final_optimization}
\begin{split} 
\mathcal{L}_{\text{MHVAE}} \coloneqq{} \frac{1}{|\mathcal{P}|} \sum_{\pi\in\mathcal{P}}\mathcal{L}^{\text{ELBO}}_{\text{MAVAE}}(x_{\pi}; \theta, \phi)  \ .
\end{split}
\end{equation}
Consequently, we propose to average all the subset-specific losses at each training iteration. The image decoding distributions are modelled as Normal with variance $\sigma$, i.e. $p_{\theta}(x_i|z_1)=\mathcal{N}(x_i;\mu_{i}(z_1), \sigma I)$, leading to reconstruction losses $-\log(p_{\theta}(x_i|z_1))$, which are proportional to $||x_i-\mu_{i}(z_1)||^{2}$. To generate sharper images, the $L_{2}$ loss is replaced by a combination of $L_{1}$ loss and GAN loss via a PatchGAN discriminator~\cite{isola2017image}. Moreover, the expected KL divergences are estimated with one sample as in \cite{maaloe2019biva}. Finally, the loss associated with the subset-specific ELBOs Eq.~\eqref{eq:final_optimization} is:
$$L=\sum_{i=1}^{N}\left(\lambda_{L_1}L_1(\mu_{i},x_{i}) + \lambda_{\text{GAN}}L_{\text{GAN}}(\mu_{i}) \right) +  \KL \ . $$
Following standard practices~\cite{isola2017image,resvit}, images are normalized in $[-1, 1]$ and the weights of the $L_1$ and GAN losses are set to $\lambda_{L_1}=100$ and  $\lambda_{\text{GAN}}=1$.

\section{Experiments}
In this section, we report experiments  conducted on the challenging problem of MR and iUS image synthesis. 
\subsubsection{Data.}
We evaluated our method on a dataset of 66 consecutive adult patients with brain gliomas who were surgically treated at the Brigham and Women's hospital, Boston USA, where both pre-operative 3D T2-SPACE and pre-dural opening intraoperative US (\iUS) reconstructed from a tracked handheld 2D probe were acquired.  The data is available at: \url{https://doi.org/10.7937/3rag-d070} \cite{remind}. 3D T2-SPACE scans were affinely registered with the pre-dura \iUS \ using NiftyReg~\cite{MODAT2010278} following the pipeline described in~\cite{drobny2018registration}. Three neurological experts manually checked registration outputs. The dataset was randomly split into a training set (N=56) and a testing set (N=10). Images were resampled to an isotropic $0.5\text{mm}$ resolution, padded for an in-plane matrix of $(192,192)$, and normalized in $[-1,1]$.

\begin{figure}[tp]
\begin{center}
\includegraphics[width=\textwidth]{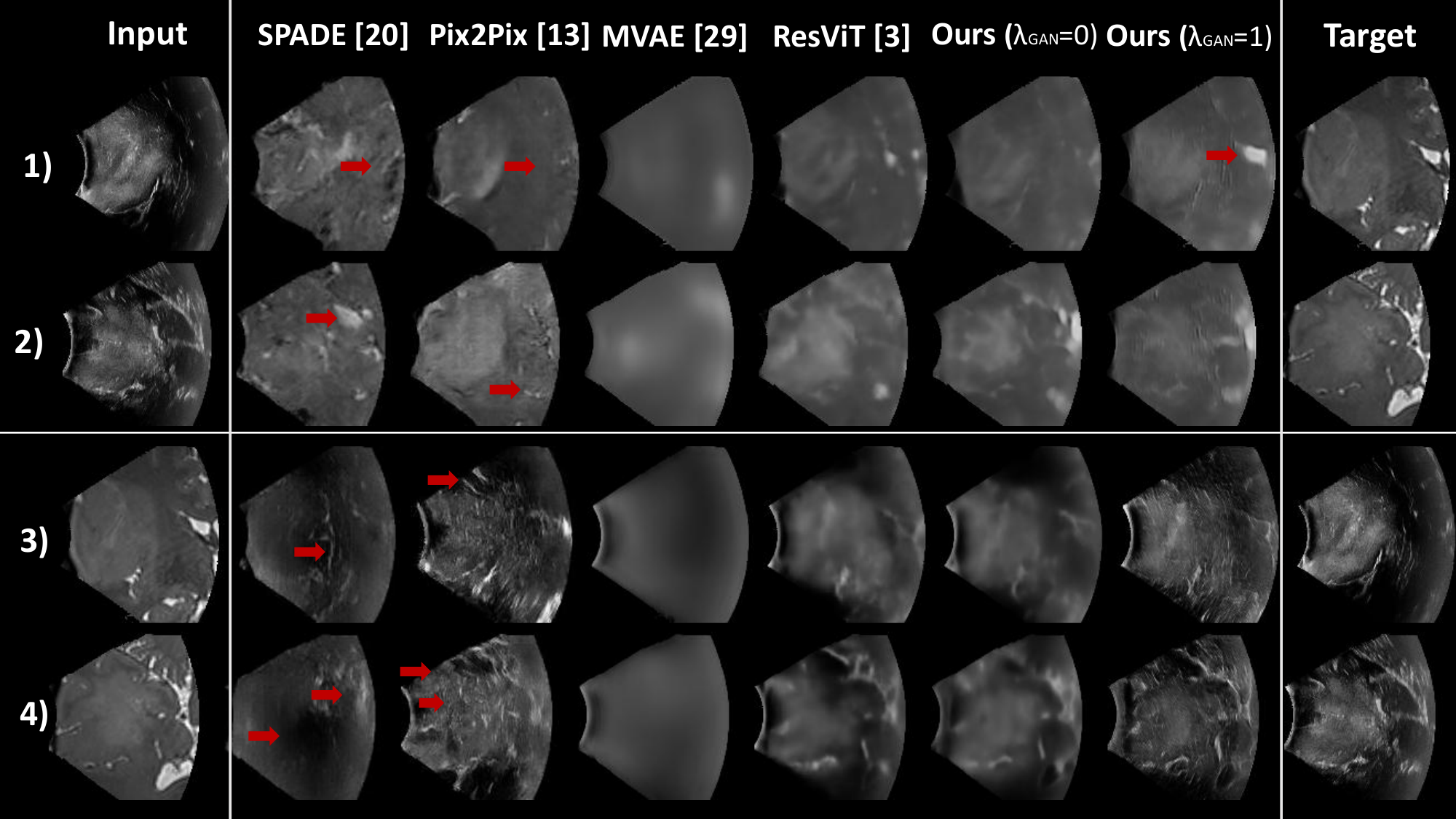}
\caption{Examples of image synthesis (rows 1 and 2: $\iUS\rightarrow\Ttwo$; rows 3 and 4: $\Ttwo\rightarrow\iUS$) using SPADE~\cite{park2019semantic}, Pix2Pix~\cite{isola2017image}, MVAE~\cite{wu2018multimodal}, ResViT~\cite{resvit} and MHVAE (ours) without and with GAN loss. As highlighted by the arrows, our approach better preserves anatomy compared to GAN-based approach and produces more realistic approach than the transformer-based approach (ResViT).} \label{fig:synthesis}
\end{center}
\end{figure}
 
\subsubsection{Implementation details.} Since raw brain ultrasound images are typically 2D, we employed a 2D U-Net-based architecture. The spatial resolution and the feature dimension of the coarsest latent variable ($z_{L}$) were set to $1\times1$ and $256$. The spatial and feature dimensions are respectively doubled and halved after each level to reach a feature representation of dimension $8$ for each pixel, i.e. $z_1\in\mathbb{R}^{196\times196\times8}$ and $z_L\in\mathbb{R}^{1\times1\times256}$. This leads to 7 latent variable levels, i.e. $L=7$.
Following state-of-the-art NVAE architecture~\cite{vahdat2020nvae}, residual cells for the encoder and decoder from  MobileNetV2~\cite{sandler2018mobilenetv2} are used with Squeeze and Excitation~\cite{hu2018squeeze} and Swish activation.  The image decoders $(\mu_i)_{i=1}^{M}$ correspond to 5 ResNet blocks. Following state-of-the-art bidirectional inference architectures~\cite{maaloe2019biva,vahdat2020nvae}, the representations extracted in the contracting path (from $x_i$ to $(z_l)_{l}$) and the expansive path (from $z_L$ to $x_i$ and  $(z_l)_{l<L}$) are partially shared. Models are trained for $1000$ epochs with a batch size of $16$. To improve convergence, $\lambda_{\text{GAN}}$ is set to 0 for the first $800$ epochs. Network architecture is presented in Appendix, and the code is available at \url{https://github.com/ReubenDo/MHVAE}.

\subsubsection{Evaluation.}  Since paired data was available for evaluation, standard supervised evaluation metrics are employed:  PSNR (Peak Signal-to-Noise Ratio), SSIM (Structural Similarity), and LPIPS~\cite{zhang2018perceptual} (Learned Perceptual Image Patch Similarity). Quantitative results are presented in Table~\ref{tab:Scores}, and qualitative results are shown in Fig.~\ref{fig:synthesis}.  Wilcoxon signed rank tests ($p < 0.01$) were performed.

\subsubsection{Ablation study.} To quantify the importance of each component of our approach, we conducted an ablation study. First, our model (MHVAE) was compared with MVAE, the non-hierarchical multi-modal VAE described in \cite{wu2018multimodal}. It can be observed in Table~\ref{tab:Scores} that MHVAE (ours) significantly outperformed MVAE. This highlights the benefits of introducing a hierarchy in the latent representation. As shown in Fig.\ref{fig:synthesis}, MVAE generated blurry images, while our approach produced sharp and detailed synthetic images. Second, the impact of the GAN loss was evaluated by comparing our model with ($\lambda_{\text{GAN}}=0$) and without ($\lambda_{\text{GAN}}=1$) the adversarial loss. Both models performed similarly in terms of evaluation metrics. However, as highlighted in Fig.~\ref{fig:synthesis}, adding the GAN loss led to more realistic textures with characteristic \iUS \ speckles on synthetic \iUS. Finally, the image similarity between the target and reconstructed images (i.e., target image used as input) was excellent, as highlighted in Table~\ref{tab:Scores}. This shows that the learned latent representations preserved the content information from input modalities. 

\subsubsection{State-of-the-art comparison.} To evaluate the performance of our model (MHVAE) against existing image synthesis frameworks, we compared it to two state-of-the-art GAN-based conditional image synthesis methods: Pix2Pix~\cite{isola2017image} and SPADE~\cite{park2019semantic}. These models have especially been used as synthesis backbones in previous MR/iUS synthesis studies \cite{donnez2021realistic,9174648}. Results in Table~\ref{tab:Scores} show that our approach statistically outperformed these GAN methods with and without adversarial learning. As shown in Fig.~\ref{fig:synthesis}, these conditional GANs produced realistic images but did not preserve the brain anatomy. Given that these models are not unified, Pix2Pix and SPADE must be trained for each synthesis direction ($\Ttwo\rightarrow\iUS$ and $\iUS\rightarrow\Ttwo$). In contrast, MHVAE is a unified approach where one model is trained for both synthesis directions, improving inference  practicality without a drop in performance. Finally, we compared our approach with ResViT \cite{resvit}, a transformer-based method that is the current state-of-the-art for unified multi-modal medical image synthesis.  Our approach outperformed or reached similar performance depending on the metric. In particular, as shown in Fig.~\ref{fig:synthesis} and in Table.~\ref{tab:Scores} for the perceptual LPIPS metric, our GAN model synthesizes images that are visually more similar to the target images. Finally, our approach demonstrates significantly lighter computational demands when compared to the current SOTA unified image synthesis framework (ResViT), both in terms of time complexity (8G MACs vs. 487G MACs) and model size (10M vs. 293M parameters). Compared to MVAEs, our hierarchical multi-modal approach only incurs a marginal increase in time complexity ($19\%$) and model size ($4\%$). Overall, this set of experiments demonstrates that variational auto-encoders with hierarchical latent representations, which offer a principled formulation for fusing multi-modal images in a shared latent representation, are effective for image synthesis.

\begin{table*}[tb]
	\centering
	\caption{Comparison against the state-of-the-art conditional GAN models for image synthesis. Available modalities are denoted by $\bullet$, the missing ones by $\circ$. Mean and standard deviation values are presented. $^{*}$ denotes significant improvement provided by a Wilcoxon test ($p < 0.01$). Arrows indicate favorable direction of each metric.
	}\label{tab:Scores}
	\resizebox{0.99\textwidth}{!}{
	\begin{tabular}{l c *{8}{c}}
		\toprule
		\multirow{2}{*}{} & \multicolumn{2}{c}{\bf Input } & \multicolumn{3}{c}{\bf \iUS} & \multicolumn{3}{c}{\bf \Ttwo }\\ 
       \cmidrule(lr){2-3} \cmidrule(lr){4-6} \cmidrule(lr){7-9}
       & \iUS & \Ttwo & PSNR(dB)$\uparrow$ & SSIM($\%$)$\uparrow$ & LPIPS($\%$)$\downarrow$  & PSNR(dB)$\uparrow$ & SSIM($\%$)$\uparrow$ & LPIPS($\%$)$\downarrow$  \\
		\midrule
\rowcolor{LightGray}
MHVAE ($\lambda_{\text{GAN}}=0$)  &  {$\bullet$}  & {$\bullet$} & 33.15 (2.48) & 91.3 (3.5) & 6.3 (2.3) & 36.38 (2.40) & 95.3 (1.9) & 2.2 (0.8)  \\
MHVAE ($\lambda_{\text{GAN}}=1$) &  {$\bullet$}  & {$\bullet$} & 31.54 (2.62) & 89.1 (4.3) & 7.1 (2.6) & 34.35 (2.67) & 93.6 (2.7) & 2.8 (1.2)\\
\midrule
\rowcolor{LightGray}
Pix2Pix \cite{isola2017image} $\Ttwo\rightarrow\iUS$  & $\circ$ & $\bullet$ & 20.31 (3.78) & 70.2 (12.0) & 19.8 (5.7) & $\times$ & $\times$ & $\times$\\ 
SPADE  \cite{park2019semantic} $\Ttwo\rightarrow\iUS$  &$\circ$ & $\bullet$ & 20.30 (3.62) & 70.1 (12.1) & 21.5 (6.9) & $\times$ & $\times$ & $\times$\\

\rowcolor{LightGray}
MVAE \cite{wu2018multimodal}  & $\circ$ & $\bullet$ & 21.21 (4.20) & 73.5 (10.9) & 26.9 (10.5) & 23.23 (4.55) & 83.4 (8.1) & 21.4 (9.0)  \\
ResViT \cite{resvit}  & $\circ$ & $\bullet$ & 21.22 (3.10) & \textbf{75.2* (9.7)} & 24.0 (7.5) & 37.14 (5.94) & 99.1 (0.9) & 1.0 (0.5) \\

\rowcolor{LightGray}
MHVAE ($\lambda_{\text{GAN}}=0$)  & $\circ$ & $\bullet$ & \textbf{21.87{*} (4.06)} & 74.9 (10.4)  & 24.2 (9.1)& 36.41 (2.13) & 95.5 (1.8) & 7.2 (3.0) \\
MHVAE ($\lambda_{\text{GAN}}=1$)  & $\circ$ & $\bullet$ & 21.26 (3.93) & 71.9 (11.4) & \textbf{19.0* (7.6)} & 34.94 (2.27) & 94.4 (2.3) & 7.6 (3.2)
\\ 
\midrule
\rowcolor{LightGray}
Pix2Pix \cite{isola2017image} $\iUS\rightarrow\Ttwo$  & $\bullet$ &$\circ$ &  $\times$ & $\times$ & $\times$ & 21.01 (3.70) & 77.9 (9.2) & 17.4 (4.7)\\ 
SPADE \cite{park2019semantic} $\iUS\rightarrow\Ttwo$  &  $\bullet$ & $\circ$ & $\times$ & $\times$ & $\times$ & 20.12 (3.20) & 74.3 (8.5) & 18.6 (3.8)\\

\rowcolor{LightGray}
MVAE \cite{wu2018multimodal}   &  $\bullet$&  $\circ$ &  23.02 (4.12) & 75.3 (10.4) & 25.5 (9.9) & 21.70 (4.60) & 82.6 (8.2) & 21.7 (9.1)\\ 

ResViT \cite{resvit}   &  $\bullet$&  $\circ$ & 35.09 (3.96) & 97.6 (1.0) & 3.5 (1.2) &21.70 (3.40) & \textbf{82.8* (7.6)} & 18.9 (6.8)\\

\rowcolor{LightGray}
MHVAE ($\lambda_{\text{GAN}}=0$)   &  $\bullet$&  $\circ$ & 33.07 (2.34) & 91.3 (3.4) & 13.2 (4.8) & \textbf{22.16* (4.13)} & \textbf{82.8* (8.0)} & 18.3 (7.6)\\ 
MHVAE ($\lambda_{\text{GAN}}=1$)   & $\bullet$ & $\circ$& 31.58 (2.26) & 90.8 (3.6) & 12.0 (4.4)& 22.12 (4.28) & 81.7 (8.2) & \textbf{17.4* (7.3))}\\ 
		\bottomrule
	\end{tabular}
	}
\end{table*}

\section{Discussion and conclusion}
\subsubsection{Other potential applications.} The current framework enables the generation of iUS data using \Ttwo \ MRI data. Since image delineation is much more efficient on MRI than on US, annotations performed on MRI could be used to train a segmentation network on pseudo-iUS data, as performed by the top-performing teams in the crossMoDA challenge \cite{dorent2023crossmoda}. For example, synthetic ultrasound images could be generated from the BraTS dataset \cite{bakas2019identifying}, the largest collection of annotated brain tumor MR scans. Qualitative results shown in Appendix demonstrate the ability of our approach to generalize well to T2 imaging from BraTS. Finally, the synthetic images could be used for improved iUS and \Ttwo \ image registration. 

\subsubsection{Conclusion and future work.} 
We introduced a multi-modal hierarchical variational  auto-encoder to perform unified MR/iUS synthesis. By approximating the true posterior using a combination of unimodal approximates and optimizing the ELBO with multi-modal and uni-modal examples, MHVAE demonstrated state-of-the-art performance on the challenging problem of iUS and MR synthesis. Future work will investigate synthesizing additional imaging modalities such as CT and other MR sequences.

\subsubsection{Acknowledgement}
This work was supported by the National Institutes of Health (R01EB032387, R01EB027134, P41EB028741, R03EB032050), the McMahon Family Brain Tumor Research Fund and by core funding from the Wellcome/EPSRC [WT203148/Z/16/Z; NS/A000049/1].
For the purpose of open access, the authors have applied a CC BY public copyright licence to any Author Accepted Manuscript version arising from this submission.

\bibliographystyle{splncs04}
\bibliography{paper2711.bib}

\newpage
\section{Appendix}
\begin{figure}[b!]
\begin{center}
\includegraphics[width=\textwidth]{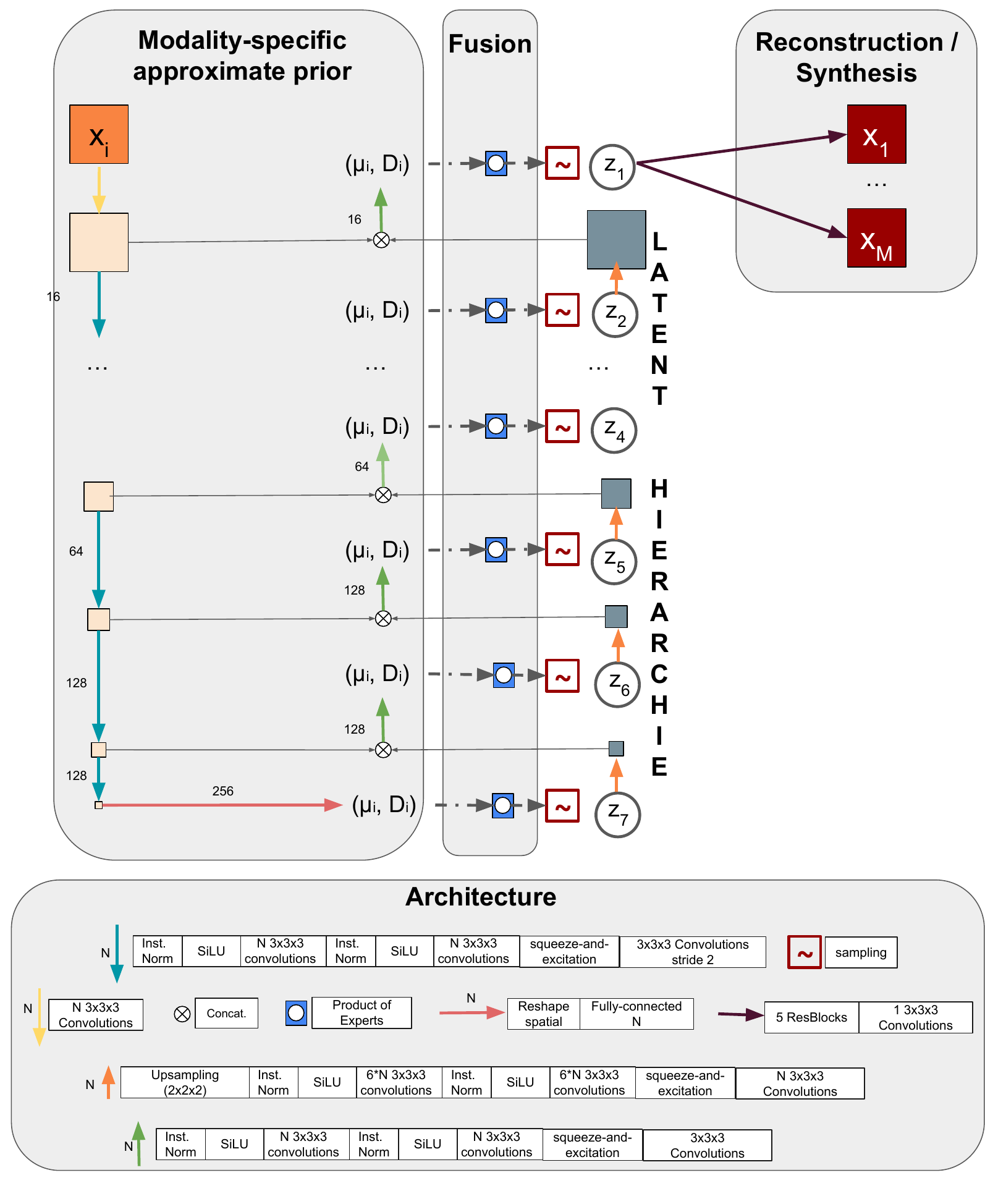}
\caption{Our multi-modal hierarchical variational auto-encoder (MHVAE). Only one modality encoder is shown. Products of experts are defined in the core manuscript.} 
\end{center}
\end{figure}

\begin{figure}[b!]
\begin{center}
\includegraphics[width=\textwidth]{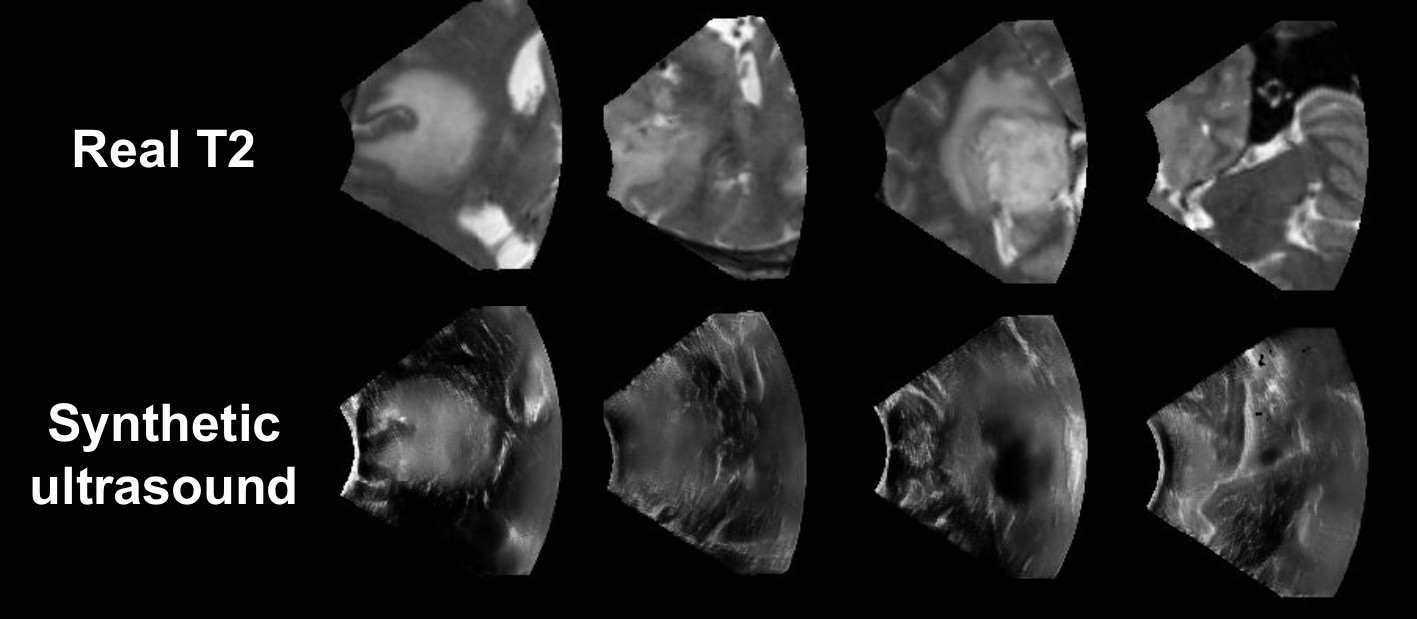}
\caption{Example of synthetic ultrasound images generated from T2 scans of the BraTS dataset.} 
\end{center}
\end{figure}

\end{document}